\documentclass[journal]{IEEEtran}
\usepackage[labelsep=period]{caption}

\ifCLASSINFOpdf
  \usepackage[pdftex]{graphicx}
  \graphicspath{{../pdf/}{../jpeg/}}
  \DeclareGraphicsExtensions{.pdf,.jpeg,.png}
\else
\fi
\ifCLASSOPTIONcompsoc
 \usepackage[caption=false,font=normalsize,labelfont=sf,textfont=sf]{subfig}
\else
 \usepackage[caption=false,font=footnotesize]{subfig}
\fi
\usepackage{amssymb}
\usepackage{xcolor}
\usepackage{amsmath}
\usepackage{subfig}
\begin{document}
% \title{Bare Demo of IEEEtran.cls\\ for IEEE Journals}
% \title{Adversarial Training for Image-to-Image Translation}
%
% \title{Explanation for Applying GAN for Image to Image Translation}
\title{Mechanisms of Generative Image-to-Image Translation Networks}

\author{Guangzong~Chen, Mingui~Sun, Zhi-Hong~Mao, Kangni~Liu, and Wenyan~Jia}
\maketitle

\begin{abstract}
% The abstract goes here.
Generative Adversarial Networks (GANs) are a class of neural networks that have been widely used in the field of image-to-image translation. In this paper, we propose a streamlined image-to-image translation network with a simpler architecture compared to existing models. We investigate the relationship between GANs and autoencoders and provide an explanation for the efficacy of employing only the GAN component for tasks involving image translation. We show that adversarial for GAN models yields results comparable to those of existing methods without additional complex loss penalties. Subsequently, we elucidate the rationale behind this phenomenon. We also incorporate experimental results to demonstrate the validity of our findings.
\end{abstract}

% \begin{IEEEkeywords}
% \textcolor{red}{IEEE, IEEEtran, journal, \LaTeX, paper, template.}
% \end{IEEEkeywords}

% \fi
\IEEEpeerreviewmaketitle

\section{Introduction}
The advancement of large neural networks has significantly improved the performance of image-to-image translation tasks. Its high accuracy and flexibility attract many researchers in various fields. Industries, such as healthcare, automotive, and entertainment, utilize image-to-image translation technologies for different applications, including medical imaging, autonomous driving, and digital content creation~\cite{Ronneberger2015,pang2022,Rombach2022}. In addition, researchers in academia and the private sectors are continuously innovating to explore new possibilities and advances in this area. Image-to-image translation encompasses a wide range of tasks, including edge-to-image, photo-to-painting, etc.~\cite{Ronneberger2015, Chen2024, Saharia2022}. All of these tasks need significant computational and data resources for the training model. Depending on the complexity of the model and the size of the dataset, training can take from hours to weeks.

A myriad of methodologies have been advanced to address the image-to-image translation problem. Despite most existing models are able to solve the problem, they do not explain the mechanisms by which the network distinguishes content from style~\cite{Zhu2017, Yi2017, Huang2018, Wang2021, Wu2019}. The nebulous definitions of content and style pose significant challenges in the mathematical characterization of the image translation process. Moreover, existing models for image-to-image translation often employ Generative Adversarial Networks (GANs) architecture, but encompass significant complexity, incorporating elements such as cycle loss, identity loss, and penalties on intermediate features. Rarely is the necessity of these intricate penalties examined.

Previously, we introduced a GAN-based model to transform food images using only GAN penalty without any additional penalties~\cite{Chen2024}. In this paper, we investigate the similarity between Generative Adversarial Networks (GANs)~\cite{Goodfellow2014} and autoencoders~\cite{Kingma2014} to elucidate the GAN model mechanism for image-to-image translation without imposing additional penalties. Subsequently, we show the rationale behind the efficacy of employing solely the GAN component for image-to-image translation tasks. We offer a clear explanation that substantiates the primary role of GAN components in addressing the image-to-image translation problem.

% \textcolor{red}{Our model and explanation.}

We have conducted a comprehensive review and analysis of the models employed for image generation and image-to-image translation. Our investigation focuses on identifying the efficacy of various components of the network. Notably, we discovered that the autoencoder and GAN models generate homologous output and provide an explanation for this phenomenon. This explanation also extends to the efficiency of GANs in the context of image-to-image translation. From our perspective, we employ a preliminary GAN for image-to-image translation. Furthermore, our findings elucidate why some examples in the network may fail.

% \textcolor{red}{We are going to explain the model from basic GAN.}

% Part 3, is there any method to overcome these disadvantages? If the answer is yes, introduce these methods shortly.

%Part 4, how our model solve the problem. What's our advantage? Could our model be applied by others easily?

  % Objective: 
This paper makes the following contributions:
(i) We demonstrate that with a discriminator of sufficient capacity to distinguish between real and synthetic images, adversarial training for autoencoder models yields results similar to those of traditional autoencoder models. This is substantiated through experimental validation.
(ii) We extend adversarial training to the image-to-image translation problem, illustrating that a straightforward GAN model can preserve common features and generate novel ones, whereas previous methods impose additional penalties to maintain common features.
(iii) Our work provides a rationale for the efficacy of GANs in the image-to-image translation context, clarifying that the decomposition of texture and content signifies common and differentiating characteristics determined by the dataset. This offers a more precise and comprehensive understanding compared to previous studies.

The paper is structured as follows: The related works section gives a brief review of image generation and translation. The methods section provides our explanation, encompassing algebraic and geometric interpretations. Subsequently, the experiment section presents three experiments. The first experiment compares the performance of GANs and autoencoders, the second investigates the model's capability for image-to-image translation, and the third examines the constraints outlined in the methods section. Finally, conclusions are drawn based on our analysis.
\section{Related Works}
% \textbf{Image Generation}
% \noindent\textbf{Generative Adversarial Networks (GANs).} \quad 
\subsection{Generative Adversarial Networks (GANs)}
GANs are widely utilized for image generation. These architectures are composed of a generator ($G$) and a discriminator ($D$) that compete in a min-max game during training. Numerous variations of GANs have been proposed to enhance their performance, such as CGAN~\cite{Odena2017, Mirza2014, Isola2017}, CVAE-GAN~\cite{Bao2017}, VQ-GAN~\cite{Esser2021}, StyleGAN~\cite{Karras2019}, GigaGAN~\cite{Kang2023} and so on~\cite{Zhou2023}. Additionally, extensive research has been conducted to address issues such as mode collapse and unstable training~\cite{Zhou2023}. These contributions substantially advance the capability of GANs in producing high-fidelity images.

% \noindent\textbf{Image translation.}\quad 
\subsection{Image Translation}
% Image generation
Gatys et al. proposed a seminal approach in which they demonstrated that style and content could be separated within a convolutional network. They used feature maps to capture the content and a Gram Matrix to capture the style~\cite{Gatys2015}. 
The style transfer has become increasingly popular with a lot of researchers. Furthermore, numerous models have been introduced for image-to-image translation. 
% style transfer
% 
CycleGAN~\cite{Zhu2017}, DualGAN~\cite{Yi2017}, and similar models posited that the transformation between two domains should be invertible. These models used two GANs to learn invertible image translation. Other approaches like MUNIT~\cite{Huang2018}, DRIT++~\cite{Lee2019}, TransferI2I~\cite{Wang2021}, assumed that style and content are controlled by different sets of latent variables. Based on this assumption, they developed various network structures to achieve the desired translations. Palette employs a diffusion model for image-to-image translation~\cite{Saharia2022}. However, its applicability is limited to tasks such as inpainting, colorization, and uncropping.

Zheng et. al.~\cite{zheng2023} addressed the issue of imbalanced image datasets using a multiadversarial framework. In addition, they introduce an asynchronous generative adversarial network to boost model performance. Yang et al. enhance the quality of the generated images through semantic cooperative shape perception~\cite{yang2024}. Additionally, researchers apply various techniques such as multi-constraints, semantic integration, and a unified circular framework to refine image-to-image translation models by modifying model specifics~\cite{Saxena2022, Li2023, Huang2022, Saxena2022, Wang2018, Wang2021b, Li2019}.

% improved style transfer -> I2I translation. But we still don't ahve clear explanation.

% \noindent\textbf{Network Explanation.}\quad
\subsection{Network Explanation}
Besides these models that provide methods for image-to-image translation, a variety of approaches have been suggested to clarify the fundamental processes driving the network's functioning from different analytical perspectives.

Classification models are essential elements of GANs. The foundational theory underlying these models is vital for the proper function of GANs. Yarotsky established error limits for network~\cite{Yarotsky2017}, while Wang et al. determined error bounds for both multi-layer perceptrons and convolutional neural networks. These studies demonstrate the theoretical correctness of convolutional neural networks~\cite{Wang2024}.

Beyond the classification model, Ye et al. introduced deep convolutional framelets as described in~\cite{Ye2018}. They utilized deep convolutional framelets to explain a model comparable to U-Net, proposing an approach that captures finer details than U-Net. This model helps to comprehend the roles of various components, such as the number of features, skip connections, and concatenation within the network.

In the context of generator networks, the variational autoencoder (VAE) and diffusion models are well explained~\cite{Kingma2014, Ho2020, Nichol2021}. The VAE focuses on minimizing the evidence lower bound (ELBO), whereas the diffusion model views the network's process as a Markov chain and derives its loss function based on the characteristics of a Markov chain. Generally, a GAN model trains a model that distinguishes the difference between real and fake. However, when GANs are applied to image-to-image translation tasks, a significant portion of the research centers on developing heuristic models, and much of the interpretation of these models is heuristic.

\section{Methods}
        \begin{figure*}[tp]
        	\centering
            \subfloat[\centering Adversarial training model]{{\includegraphics[width=11cm]{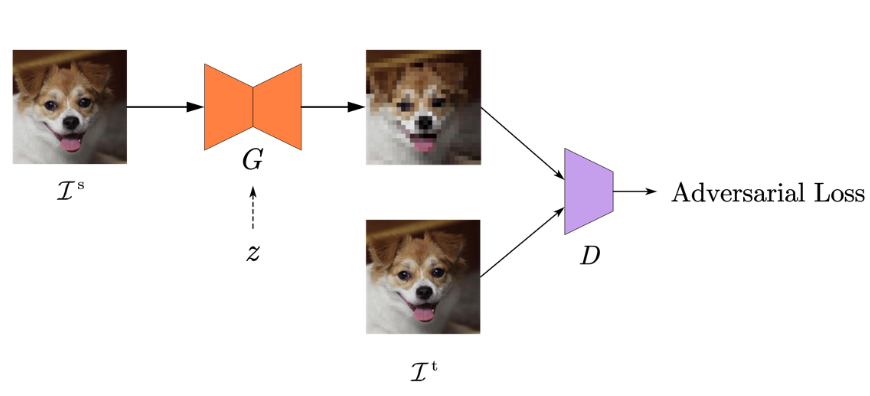} }}%
            \qquad
            \subfloat[\centering Autoencoder model]{{\includegraphics[width=6cm]{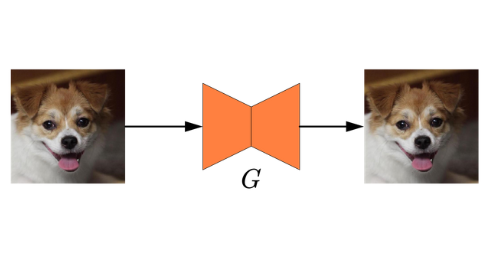} }}%
        	\caption{The architecture of the method.}
        	\label{img:overview}
        \end{figure*}

% \textcolor{red}{Define what is adversarial training and autoencoder in this section.}

    The aim of this section is to elucidate the mechanism of adversarial training within the context of image-to-image translation challenges. Initially, we focus on a specific instance: the identity image translation task. Subsequently, we broaden our analysis to encompass the general image-to-image translation paradigm, providing a comprehensive explanation to demonstrate how GAN models can be applied to image-to-image translation tasks.

    The task of recovering an image from a latent space is commonly addressed through autoencoders. This issue is similar to the image reconstruction. However, in image reconstruction, the input image may exhibit certain defects that require correction. In contrast, in our scenario, the input and output images are identical. Our findings demonstrate that employing either of the two methodologies yields similar results. Consequently, these conclusions can be extrapolated to the image translation problem.
    
    Autoencoders are widely employed to derive latent variables from input images. It is also used in image reconstruction applications. The main objective of an autoencoder is to learn a mapping function $G(x)$, capable of reconstructing the input image $x$. The generator $G$,  comprises an encoder and a decoder, where the encoder is utilized to obtain the latent variable and the decoder reconstructs the image from the latent variable.
    
    Adversarial training, in this paper, is defined by the introduction of a mapping function $D$ which apparent the differences between authentic images $x$ and reconstructed images $G(x)$. It is similar to the discriminator function in GAN. The difference between the mapping function $D$ and the discriminator in GAN is that $D$ does not use binary output while the discriminator function in GAN requires binary output. The discriminator in GANs is a special case of the mapping function $D$.
    The training framework is a min-max game between $G$ and $D$, in which $D$ aims to maximize the loss function, while $G$ aims to minimize it. 
    
    % Adversarial training differs from Generative Adversarial Networks (GANs) in that it does not require a binary output from the discriminator $D$. However, we will show that the results of adversarial training and GANs are similar when the discriminator possesses the adequate capability to distinguish the difference between $x$ and $G(x)$.
    
    % \textcolor{red}{Insert Fig. 1 to explain the model.}
   Fig.~\ref{img:overview} shows two distinct network architectures for generative learning and autoencoder. The right is adversarial training. The left is the autoencoder. For the autoencoder, the goal is to employ a model to recreate the input data. Adversarial training involves alternating the learning of $G$ and $D$, where $G$ generates images and $D$ identifies the differences between the input and the generated output.
   A random variable $z$ is sampled from a Gaussian distribution and used exclusively to produce multiple outputs from a single input image. The image datasets $\mathcal{I}^\text{s}$ and $\mathcal{I}^{\text{t}}$ represent distinct datasets, where $\mathcal{I}^\text{s}$ is used as shape references and $\mathcal{I}^\text{t}$ is used to provide texture information. When comparing the autoencoder with adversarial training, we set $\mathcal{I}^{\text{s}}$ and $\mathcal{I}^{\text{t}}$ to be identical.
   % \textcolor{red}{This may not accurate.}
    
% \subsection{Identical Image Generation}
% \subsection{Self-Referential Image Generation}
\subsection{Similarity Between Autoencoder and Adversarial Training Under Certain Condition}
  In this subsection, we demonstrate that autoencoders and adversarial training yield similar results given two specific constraints. Firstly, the generator must have the ability to reconstruct the input image. Secondly, the mapping function $D$ should accurately perceive the distinction between $x$ and $G(x)$.

    % We provide two explanations for the results similarity between adversarial training and autoencoder: Algebraic and geometric. % adj or n? 

    \subsubsection{Algebraic Explanation}

        Let $\mathcal{I} = \{x^{(1)}, x^{(2)}, ..., x^{(m)}\}$ be a set of data, where $x^{(i)} = \left[ x^{(i)}_1, x^{(i)}_2, \dots, x^{(i)}_n \right]^T \in\mathbb{R}^n$.

        The optimization problem of the autoencoder is formulated as:
        \begin{equation}
            \min_G L = \frac{1}{m} \sum_{x \in \mathcal{I}} ||x - G(x)||
        \end{equation}
        where $\|\cdot\|$ is $L_1$ norm, which is the sum of the absolute values of the element of vector.
        
        The adversarial training incorporates an additional mapping function $D$, which maps $x$ to a vector $D(x)$, with $D(x)$ belonging to $\mathbb{R}^n$.After transformation, in the new space, $D(x)$ and $D(G(x))$ are linearly separable. It is important to note that the GAN requires a binary output from the discriminator, whereas the mapping function $D$ projects to a new space with dimension $n$.
        
        The optimization problem of adversarial training is defined as follows:
        
        \begin{equation}
            \min_G \max_D L =  \frac{1}{m} \sum_{x \in \mathcal{I}} \|D(x) - D(G(x))\|
            \label{eq:adv}
        \end{equation}
        where $D(x)$ is
        $ D(x) = \left[  D_1(x), D_2(x),\cdots, D_n(x) \right]^\text{T}.$

        The main difference between autoencoders and adversarial training is the presence of an auxiliary function $D$. This additional component augments the differences between the input data points $x$ and their generated data $G(x)$, which helps to train the generator. Both algorithms aim to make $G(x)$ close to $x$, leading to similar results. However, they might produce different results because, near the optimal solution, $D$ in adversarial training can become oscillating, causing $G$ to fluctuate around the optimum. In contrast, the autoencoder will converge to the optimal solution.

        In (\ref{eq:adv}), the training data is paired, which means $x$ and $G(x)$ must be considered together when computing the loss function. We will now demonstrate that adversarial training can be performed without paired data. If the function $D$ can maximize the loss function and perfect distinguish between $x$ and $G(X)$ on each feature, then there must be a function $D$ that $D_i(x) > D_i(G(x))$ and optimize the loss function at the same time. Then we have the following loss function:
        
        % \begin{equation}
        %     L = \frac{1}{m} \sum_{x \in I} \|D(x) - D(G(x))\|.
        % \end{equation}
        
        \begin{equation}
            L = \frac{1}{m} \sum_{x\in \mathcal{I}}\sum_{i} [D_i(x) -  D_i(G(x))].
        \end{equation}
        Rearranging the equation, we have:
        \begin{equation}
            \begin{aligned}
                L  &= \frac{1}{m} \left[\sum_{x \in \mathcal{I}} \sum_{i} D_i(x) - \sum_{x \in \mathcal{I}}  \sum_{i} D_i(G(x)) \right]. \\
                % &= \frac{1}{m} \sum_{x \in I} D(x) - \sum_{x \in I} D(G(x)). \\
            \end{aligned}
        \end{equation}

        We can define another function $\hat{D}(x) = \sum_{i} D_i(x)$,   where $\hat{D}(x) \in \mathbb{R}$. And the loss function can be written as:
        \begin{equation}
            \begin{aligned}
                L &= \frac{1}{m} \left[\sum_{x \in \mathcal{I}} \hat{D}(x) - \sum_{x \in \mathcal{I}} \hat{D}(G(x))\right].
            \end{aligned}
        \end{equation}

        Because $D$ only required to distinguish different features in $x$ and $G(x)$, we consider using random variables and distribution to model the problem. Let $p_{\text{data}}$ be the distribution of the data set and $p_{\text{g}}$ be the distribution of the generator's output, and replacing average with expectation, then we have
        \begin{equation}
            \begin{aligned}
                L &= \mathbb{E}_{x \sim p_{\text{data}}(x)} [\hat{D}(x)] - \mathbb{E}_{x \sim p_\text{g}(x)} [\hat{D}(x)].
            \end{aligned}
            \label{eq:adv_final}
        \end{equation}
        This is similar to the WGAN loss function~\cite{Arjovsky2017}.
        From (\ref{eq:adv}), we know that $G(x)$ will be push to $x$ when minimizing the loss function. Therefore, adversarial training should produce results similar to autoencoder models. The equation (\ref{eq:adv_final}) tells us that if the discriminator $D$ can perfectly distinguish the data from $p_{\text{data}}$ and and $p_{\text{g}}$, the loss function will not depend on the order of $x$ and $G(x)$. 

        % \begin{algorithm}
        %     \caption{Adversarial Training}
        %     \label{alg:adv}
        %     \begin{algorithmic}
        %         \Require $\Theta_G$, $\Theta_D$; (parameter of $G$ and $D$).
        %         \State Initialize $G$ and $D$ with parameters $\Theta_G$ and $\Theta_D$.
        %         \For[ $i = 1$ to $epochs$]
        %             \State Sample a batch of real data $x$ from the data distribution
        %             \State Generate a batch of reconstructed data $x' = G(x)$

        %             \State \textbf{Train the $D$:}
        %             % \State $L = \mathbb{E}[||D(x) - D(x')||_2^2] - \lambda ||\Theta_D||_2$
        %             \State $L = \mathbb{E}[D(x)] - \mathbb{E}[D(x')] - \lambda ||\Theta_D||_2$
        %             \State $\Theta_D \leftarrow \Theta_D + \eta_D \nabla_D L$
        %             % \State Update the discriminator parameters $\Theta_D$ to maximize $L_D$

        %             \State Sample a batch of real data $x$ from the data distribution
        %             \State Generate a batch of reconstructed data $x' = G(x)$
        %             \State \textbf{Train the generator:}
        %             \State $L = \mathbb{E}[D(x)] - \mathbb{E}[D(x')]$
        %             \State $\Theta_G = \Theta_G - \eta_G \nabla_{\Theta_G} L$
        %             % \State Compute the loss $L_G = \mathbb{E}[||D(G(x)) - D(x)||_2] - \lambda ||\Theta_G||_2$
        %             % \State Update the generator parameters $\Theta_G$ to minimize $L_G$
        %         \EndFor
        %     \end{algorithmic}
        % \end{algorithm}

    \subsubsection{Geometric Interpretation}
        % \begin{figure*}[tp]
        % 	\centering
        % 	\includegraphics[height=5.5cm]{imgs/GE.png}
        % 	\caption{Geometric interpretation. The left shows the data in their original form. The mapping function $D$ transforms the data into a different space, and a linear boundary can be created in a new space.}
        % 	\label{img:geoexp}
        % \end{figure*}
        \begin{figure*}
            \centering
            \includegraphics[width=0.85\linewidth]{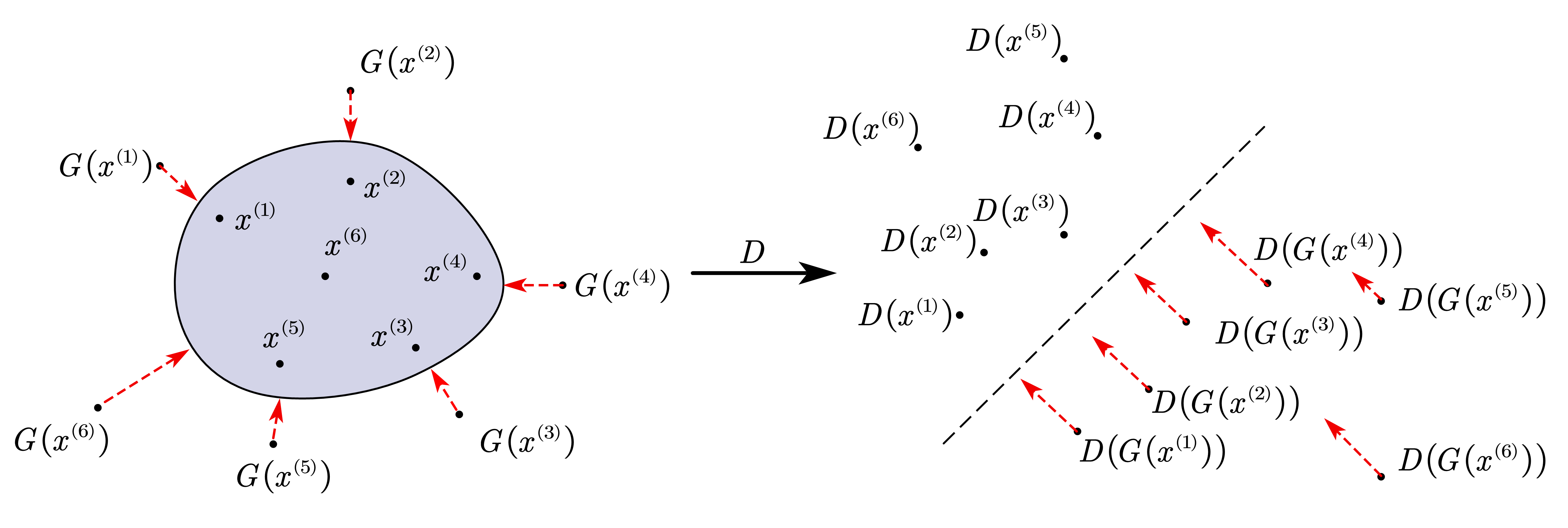}
            \caption{Geometric representation of initial phase of the model.}% In the begin, $D(x^{(i)})$ and $D(G(x^{(i)})$ can be separate easily in $D$ space. }% 
            %左右两个部分分别代表什么？中间的D转换是什么？需要说明清楚%
            \label{fig:ge_init}
        \end{figure*}
        \begin{figure*}
            \centering
            \includegraphics[width=0.85\linewidth]{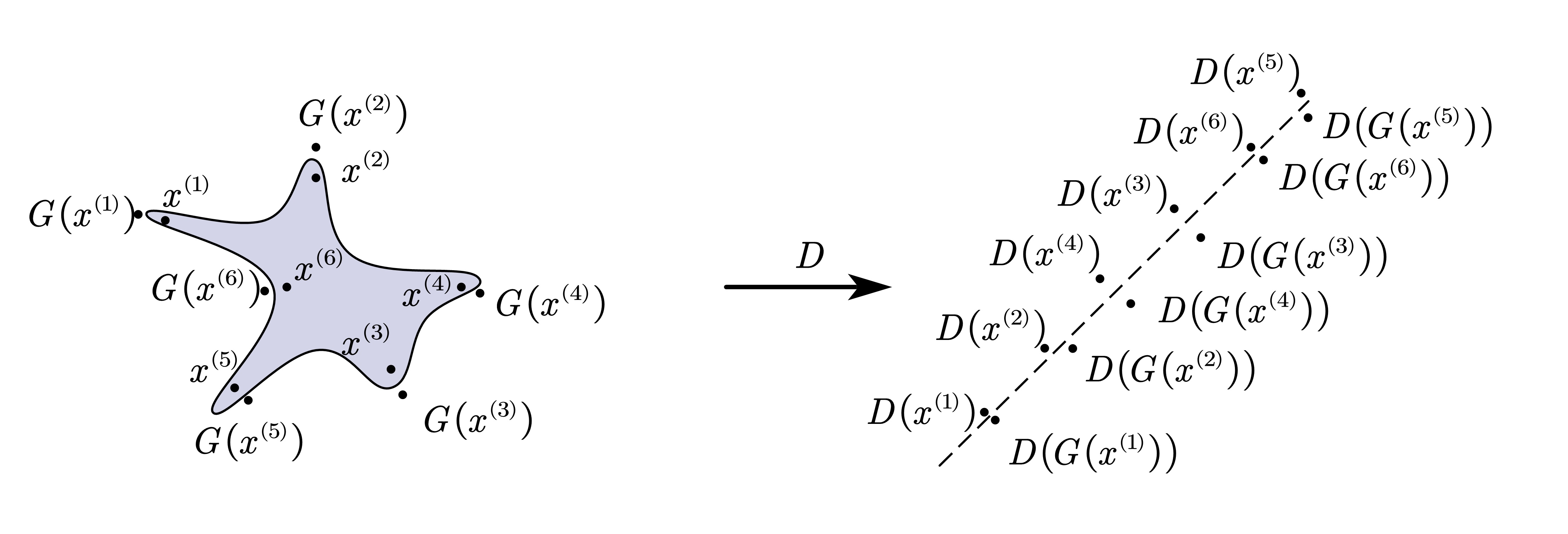}
            \caption{Geometric representation of the model after alternating training $G$ and $D$.}%同上%
            \label{fig:ge_res}
        \end{figure*}
        We also present a geometric interpretation of why adversarial training can produce results similar to the autoencoder. 
        Fig.~\ref{fig:ge_init} shows the status of the early stage of the model training.
        After learning $D(\cdot)$ in the max part of the min-max optimization problem (\ref{eq:adv}), we project $x$'s and $G(x)$'s onto a new feature space where the set of $x$'s and the set of $G(x)$'s are well clustered and can be separated by a hyperplane---a linear boundary, similar to how the data with different labels are separated in the support vector machine (SVM).
        If we map the dividing surface to the original space, a nonlinear boundary will emerge to distinguish $x$'s from $G(x)$'s.
        % For every pair of points $x$ and $G(x)$, the boundary is perpendicular to the line segment that connects $x$ and $G(x)$.
        When solving the min part of the min-max optimization problem for $G(\cdot)$, $G(x)$'s will move toward the boundary, getting closer to $x$'s, as demonstrated by the red arrows in Fig.~\ref{fig:ge_init}.
        % which is shown as the red arrow in Fig.~\ref{fig:ge_init}.
        Through the alternating training of $G$ and $D$, $G(x)$'s become closer to the set of $x$, effectively pushing both $G(x)$'s and $x$'s toward the boundary. This process is likely to bring each pair of $G(x)$ and $x$ close to each other.
        % and thus close to $x$.
       
    Fig.~\ref{fig:ge_res} illustrates the effects of $G(\cdot)$ and $D(\cdot)$ after training. Within the transformed space, $D(x)$'s and $D(G(x))$'s are distributed along the hyperplane. In the original space, the boundary is nonlinear, and $x$'s and $G(x)$'s scatter close to each other.
    
    From this perspective, the result of adversarial training will be similar to the autoencoder when $\mathcal{I}^{\text{s}}$ and $\mathcal{I}^{\text{t}}$ are from the same distribution.
    This observation may contradict our initial expectations that GANs could generate any sample that fits the distribution of the dataset. However, our findings indicate that the adversarial model will produce the input data without imposing a reconstruction penalty between $x$ and $G(x)$.

        % The boundary is solving the minimum part of the optimization problem for $G(\cdot)$ will result in $G(x)$ close to the dataset of $x$, effectively bringing $G(x)$ close to the boundary and therefore near $x$. 
        
        % \textcolor{red}{This process is show in the initial state when training $G$. All $G(X)$ are training to close to the boundary.
        
    \subsection{Image-to-Image Translation}

    The network architecture is depicted on the left of Fig.~\ref{img:overview}. It incorporates two datasets: The first image dataset, $\mathcal{I}^{\text{s}}$, is used as shape reference
    , where $\mathcal{I}^{\text{s}}$ equals $\{I^{\text{s}}_i\,|\, i = 1,\dots, N\}$,  $I^{\text{s}}_i \in \mathbb{R}^{ H \times W \times 3}$, $H$ and $W$ are the height and width of the images, $3$ is the number of channels of an RGB image, and $N$ is the total number of images. The second image dataset, $\mathcal{I}^\text{t}$, is used to provide texture information, where $\mathcal{I}^{\text{t}}$ equals $\{I^{\text{t}}_i\,|\, i = 1,\dots, M\}$, $I^{\text{t}}_i \in \mathbb{R}^{ H \times W \times 3}$, and $M$ the size of the second dataset.
    This dataset is provided to the discriminator $D$, to train the network. 
    
    We want to apply $\mathcal{I}^{\text{s}}$ to facilitate the network to generate images with the same shapes as the images in $\mathcal{I}^{\text{s}}$ while maintaining the textures from $\mathcal{I}^{\text{t}}$.
    % Ensure that the generated image conforms to the $\mathcal{I}^{\text{t}}$ distribution while maintaining the shape of the input image.
    For example, zebras and horses share a common body shape but differ in texture. The dataset $\mathcal{I}^{\text{s}}$ comprises horse images, whereas the $\mathcal{I}^{\text{t}}$ consists of zebra images. Image translations aims at substitute the horse image texture with that of the zebra.

    In the self-translation task, the mapping function $D$ is required to verify that all features in both $x$ and $G(x)$ are identical. On the other hand, in the image-to-image translation task, the discriminator's role is to confirm that all features in the generated image match the distribution of the $\mathcal{I}^{\text{t}}$.

    Consider an input image with two feature sets, $x = [x_1, x_2]$, where $x_1$ appears in both $\mathcal{I}^{\text{s}}$ and $\mathcal{I}^{\text{t}}$, but $x_2$ is only found in $\mathcal{I}^{\text{t}}$. In this case, the network will preserve the feature $x_1$ and substitute $x_2$ with a feature from the $\mathcal{I}^{\text{t}}$ dataset. The preservation of $x_1$ was explained in the previous section. Adversarial training will maintain all features if they are presented in the $\mathcal{I}^{\text{t}}$ dataset.

\section{Experimental Results and Discussions}

We conducted three experiments. First, we verified our theoretical finding that GANs produce similar results with the autoencoder models when the reference images $\mathcal{I}^{\text{s}}$ and texture images $\mathcal{I}^{\text{t}}$ are the same. Second, we showcased the GAN model's capability to transform images from one domain to another. Lastly, we modified the dataset size and the generator configuration to examine the impact of the constraints as discussed in the Methods section.
%% \href{https://blog.ggeta.com/post/Image\%20to\%20image\%20translation}{link} \textcolor{red}{Add some overview of our experiments.}
%The experiments are organized in three parts. In the first part, we show the network has
%ability to transfer the image from one domain to another domain. It can
%generate comparable results with other methods. In the second part, we show that the
%GAN can get the same result as an autoencoder because of the network structure.
%In the third part, we show what the network can generate after initialization.
%\textbf{Feature analysis of images}.

% \noindent\textbf{Datasets}\quad
We evaluated our model on various datasets, such as Animal FacesHQ (AFHQ)~\cite{Choi2017}, Photo-to-Van Gogh, Photo-to-Monet from CycleGAN~\cite{Zhu2017}, and Flickr-Faces-HQ (FFHQ)~\cite{Karras2019}.
The AFHQ dataset consists of $16130$ images of animal faces, each with a $1024 \times 1024$ pixel resolution, covering three categories of animals: cat, dog, and wild. 
The Photo-to-Van Gogh and Photo-to-Monet has approximately 1000 images for each category. The FFHQ dataset is a high-quality collection of human facial images. It comprises $70000$ images, all at a resolution of $1024 \times 1024$ pixels. In this study, we resized images to $512 \times 512$ resolution for all experiments. 

% \noindent\textbf{Implementation Details}\quad
For both the generator and discriminator, we utilized StyleGAN v2~\cite{Karras2019} as the foundational architecture. Given that an additional encoder is required to encode the image into features, we used a simple convolutional network as the encoder, which comprises only convolution, downsampling, and ReLU activation.

\subsection{Comparison Between GAN and Autoencoder}
In this subsection, we used the AFHQ dataset to demonstrate the correctness of our analysis in the methods section.
        \begin{figure}[tp]
        	\centering
        	\includegraphics[height=7cm]{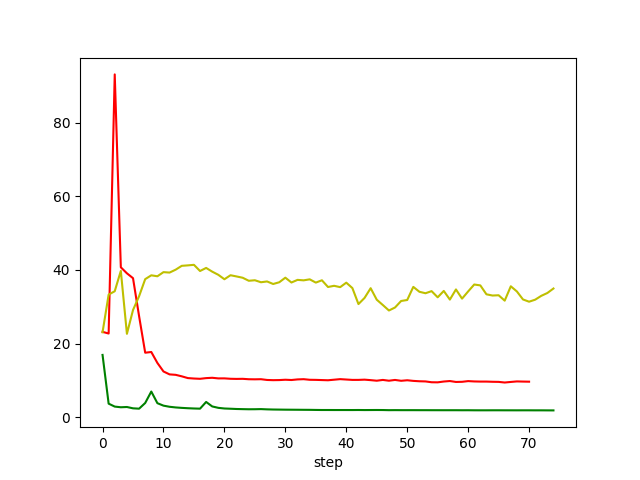}
        	\caption{Reconstruction losses from three distinct training sessions. Green: Autoencoder; Red: GAN; Yellow: GAN for image-to-image translation.}%图片的横轴和纵轴的数字太小了，放大，加粗%
        	\label{fig:rec_loss}
        \end{figure}

We claim that GANs and autoencoders can produce similar results when the generator and discriminator have enough capacity. We used the mean square error between orignal image and generated image to evaluate the performance of the two models. Fig.~\ref{fig:rec_loss} shows the reconstruction loss for three different models during the training phase.

To ensure a fair comparison between the GAN and the autoencoder, we computed the reconstruction loss for generations after every $1000$ images used to train the model. The green curve is generated by the autoencoder, the red curve by the GAN, and the yellow curve represents the reconstruction loss of the GAN model, when $\mathcal{I}^{\text{s}}$ and $\mathcal{I}^{\text{t}}$ differ.
These findings suggest that when $\mathcal{I}^{\text{s}}$ and $\mathcal{I}^{\text{t}}$ are equivalent, both the GAN and autoencoder are effective in minimizing the reconstruction loss. Despite the fact that reconstruction loss is not utilized during the training of the GAN model, this reinforces the validity of our analysis in methods section.

% The figure \ref{fig:rec_loss} show the reconstruction losses, which is \(||x - G(x)||\).  The blue line is the reconstruction loss of GAN model. The reconstruction loss is used to show measure whether adversarial  model has similar result as autoencoder. It does not used to train the model. 
% \ref{fig:AFHQ_GAN} show the progress of generator result with the same input.
% From this image, that we can
        \begin{figure*}[tp]
        	\centering
        	\includegraphics[width=16cm]{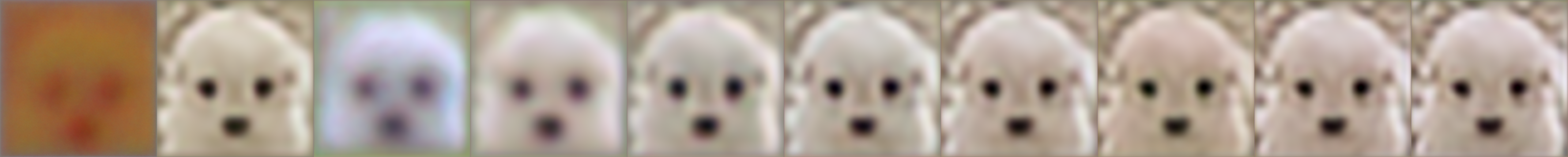}
            \includegraphics[width=16cm]{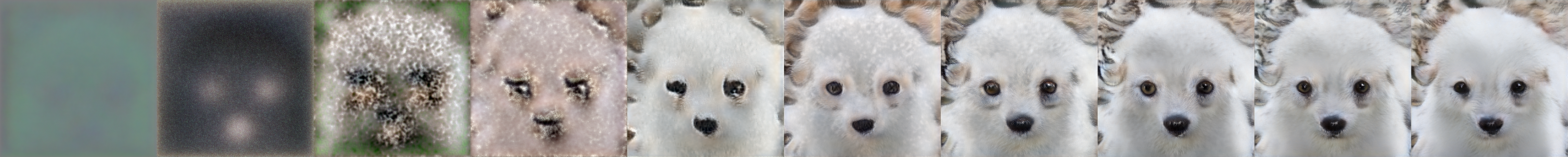}
        	\caption{Intermediate results from the autoencoder and GAN, with the top row from the autoencoder, and the bottom row from the GAN.}%
        	\label{fig:AFHQ_PROGRESS}
        \end{figure*}
Fig.~\ref{fig:AFHQ_PROGRESS} shows the outputs from the generator. This illustration makes it clear that the discriminator network starts by focusing on global features and then transitions to focusing on local features~(first row of Fig.~\ref{fig:AFHQ_PROGRESS}). In contrast, the autoencoder behaves differently, as it directly minimizes the loss across the entire image~(second row of Fig.~\ref{fig:AFHQ_PROGRESS}).

        \begin{figure*}[tp]
        	\centering
        	\includegraphics[width=16cm]{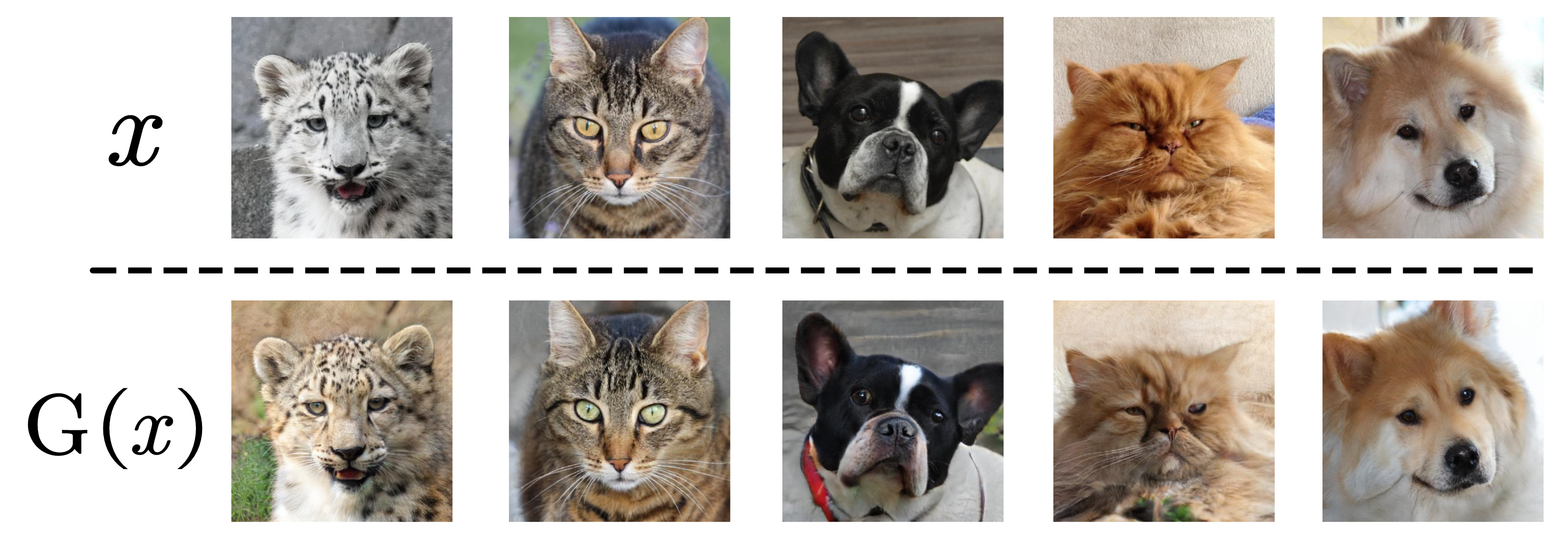}
        	\caption{Input and generated images. The top row displays the original images, while the bottom row is the generated images.}%什么模型generate的图片？需要说明%
        	\label{fig: AFHQ_ORI_FINAL}
        \end{figure*}
Fig.~\ref{fig: AFHQ_ORI_FINAL} shows both the original images and the generator's outputs. This result indicates that the outputs of the generator is similar to the input images. However, there are noticeable differences between the input and output images, such as variations in color and background. This result also illustrates the gap between the GAN and autoencoder in Fig.~\ref{fig:rec_loss}. The GAN is capable of bringing $G(x)$ close to $x$, but it cannot make them identical without incorporating a reconstruction loss.

\subsection{Image-to-Image Translation Capability}

\begin{figure*}[ht]
	\centering
	\includegraphics[width=12cm]{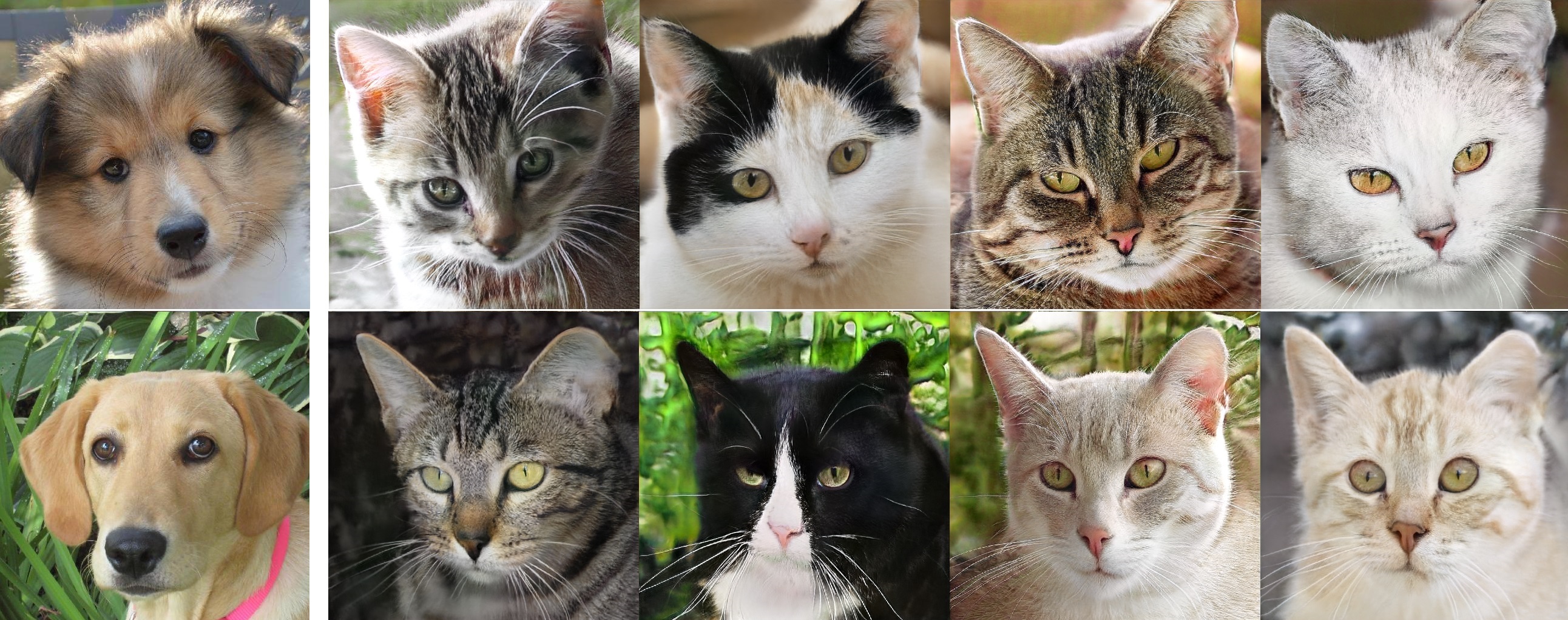}
	\caption{Results of animal image translation. First column is the input images and rest are generated images.}%说明是哪个模型generate的图片，谢谢。另外，Fig7是5列，Fig8-12都是9列。是否需要统一格式？要不然Fig8-12里面的每一张图看上去都太小了%
	\label{fig:cat-dog}
\end{figure*}

\begin{figure*}
	\centering
	\includegraphics[width=16cm]{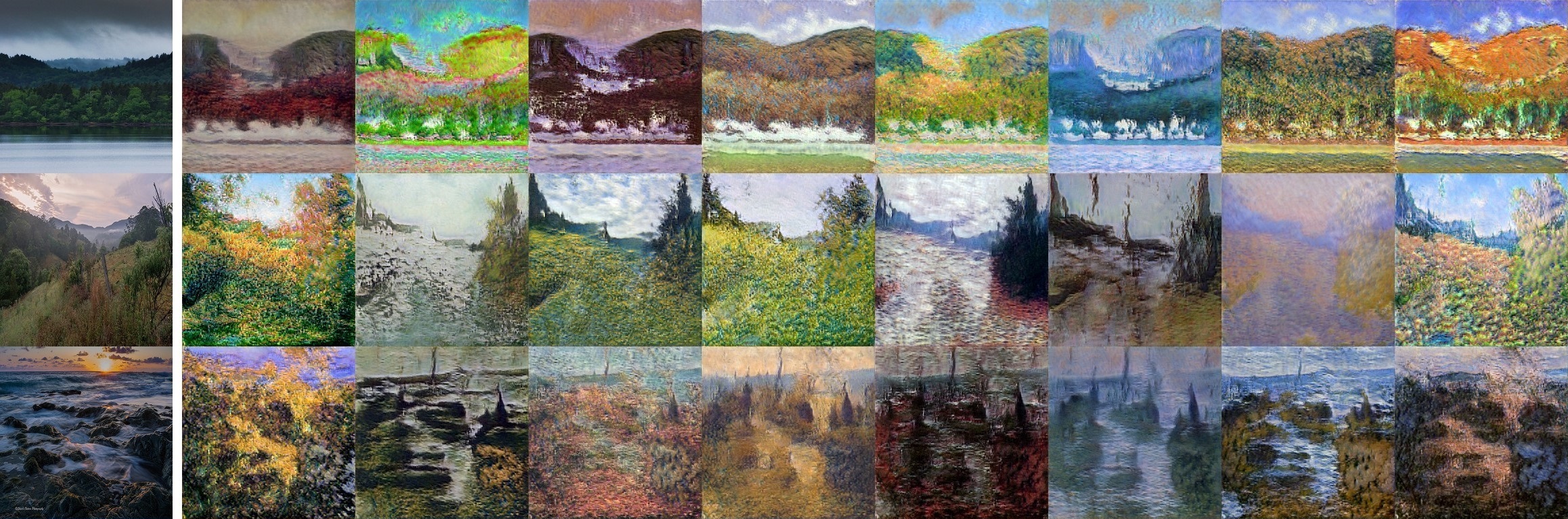}
	\caption{Translation from photo to Monet style. First column is the input image, and rest are generated images.}%同上%
	\label{fig:Style-Monet}
\end{figure*}
When $\mathcal{I}^{\text{s}}$ and $\mathcal{I}^{\text{t}}$ are different, our method can be used for image-to-image translation and the same feature in both dataset will be preserved. Compared to other methods, the network is simpler, and we can predict the outcomes and provide explanations for the results.

% \noindent\textbf{Encoder structure. \ } The encoder used in this experiment is a simple convolutional network. It first downsamples to $3 \times 128 \times 128$, then goes through $3$ convolutional blocks. Each block contains a convolutional layer, a ReLU activation layer, and a downsampling layer. The number of intermediate channels is $128$. The dimension of the intermeidate feature is $16 \times 16 \times 128$.

% \noindent\textbf{Results.}
Fig.~\ref{fig:cat-dog} shows animal transfer examples. The first column displays the input, followed by four columns showing the outputs. The dogs faces are used as $\mathcal{I}^{\text{s}}$ and cats faces as $\mathcal{I}^{\text{t}}$. The generated cat face retains the same orientation as the dog face. In addition, the relative positions of facial features such as the eyes, nose, and ears remain uniform.

A translation between an artwork and a photograph is also illustrated. In Fig.~\ref{fig:Style-Monet}, the first column shows the input, while the subsequent columns show the output. It is noticeable that the objects remain the same, but the textures are different in the output. However, in the first row, the shape of the mountain appears slightly altered. According to our explanation, this happens because the input shape of the mountain is absent in the target dataset, causing the network to modify the mountain's shape. 
% \textcolor{red}{objects(mountain, tree, lake) in the photo with different Monet painting style. This is incorrect.}

In both animal and artwork translations, it shows successes in preserving global topological characteristics. The results of these experiments show that our network can have similar results to other style transfer networks.

From this experiment, we can roughly tell what the shape (content) and style are in other style transfer models, while the other models did not explicitly indicate the content and style. In the AFHQ dataset, the style may refer to the breed of the animal, and the content refers to the pose and angle of the animal. In the Photo-to-Van Gogh dataset, the style refers to the color and texture of the picture, and the content refers to the objects in the picture. However, in our work, we can tell that the network does not have a semantic understanding of the image. The content actually refers to the common features in both datasets, and the style refers to the features only present in $\mathcal{I}^{\text{t}}$ but not in $\mathcal{I}^{\text{s}}$.
% The content actually is the low-frequency feature and the style is the feature other than global topology features which will include high frequency features and details in the image.

\subsection{Constraints Analysis}

% Our results in last section are rely on the generator is able to reconstruct the input image, which it should satisfies the perfect reconstruction condition\cite{}.
In the previous subsection, we demonstrated that our method can generate results similar to those of an autoencoder and also shows that the network has the capacity to solve image-to-image translation tasks. However, our method hinges on two critical conditions: first, the generator must be capable of completely reconstructing the input image; second, the discriminator must be able to perfectly distinguish between real and fake images whenever there is a discrepancy. In this subsection, we discuss the impact of these two conditions. We consider the generator to be composed of two parts: an encoder and a decoder. If the encoder's capacity is insufficient, it can only retain certain features, implying that the generator will fail to produce an exact match of the input when dealing with a large dataset. Evaluating the condition on discriminator is inherently challenging, but it is known that smaller dataset makes it easier for the network to memorize the entire dataset. Therefore, we present results based on various dataset sizes.

We conducted two experiments to illustrate how the abovementioned two conditions influence the network performance. We employed both the FFHQ and AFHQ datasets, as they allow us to compare the effects of dataset size. We utilizing varying sizes of intermediate features. Employing smaller intermediate features results in increased difficulty in reconstructing the input image. We find that the network initially captures the global topological features, followed by the detailed ones, which is the same as we observed in the first experiment. If the size of the dataset is sufficiently small, which means that the network has the ability to distinguish between $G(x)$ and $x$, the network tends to converge towards a one-to-one mapping.

% \noindent\textbf{Encoder structure.\ }
% Next we discuss about the encoder structure.
The first experiment kept the same encoder structure as in the previous experiment, where the intermediate feature is $16 \times 16 \times 128$ referring as \textit{high dimension feature}. In the second experiment, we add one more convolutional block in the encoder. The intermediate feature comes to $8 \times 8 \times 128$ which refer as \textit{low dimension features}. In low dimension feature, the encoder makes the information more campact, and more information are lost.

\subsubsection{Image Translation With High Dimension Features}

% Except translation between 2 groups of items, self translation has also implemented in our network. A face to face translation is realized and the result is shown in Fig.~\ref{fig:f2f16x16x128}.  The FFHQ dataset contains \textcolor{red{70,000} human face photos but AFHQ dataset only have \textcolor{red{3,000} animal face photos.

\begin{figure*}
	\centering
	\includegraphics[width=16cm]{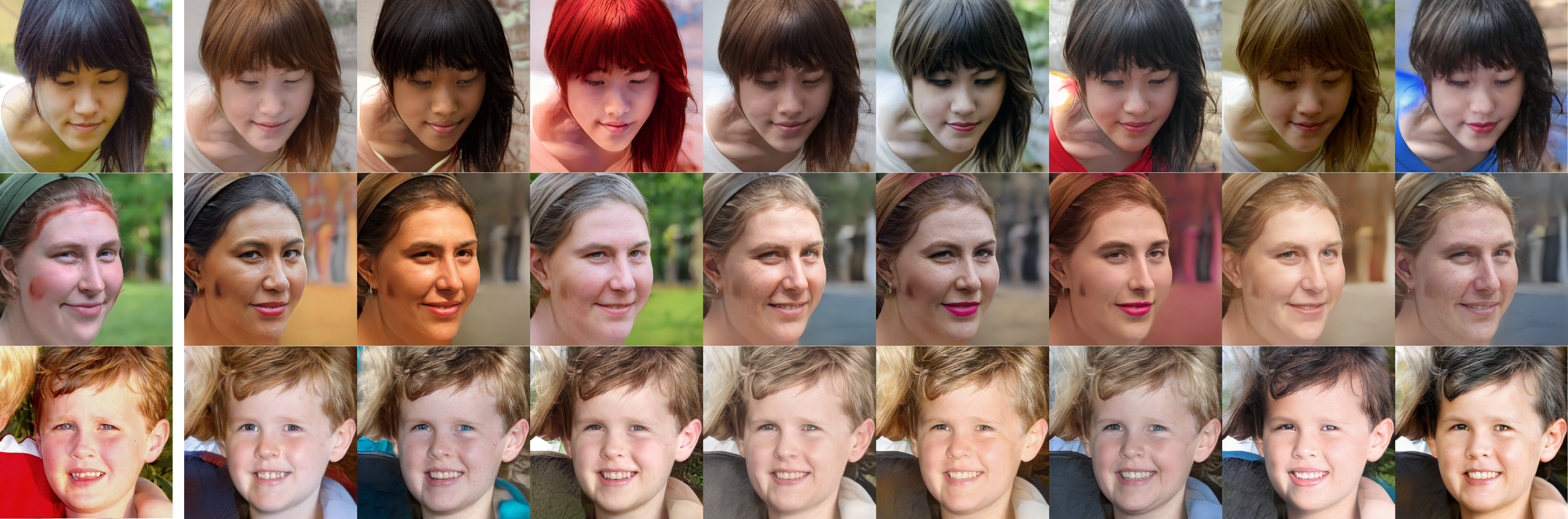}
	\caption{Face-to-face translation results with $16\times 16 \times 128$ intermediate features. First column is the input images, and rest are generated images.}
	\label{fig:f2f16x16x128}
\end{figure*}

\begin{figure*}
	\centering
	\includegraphics[width=16cm]{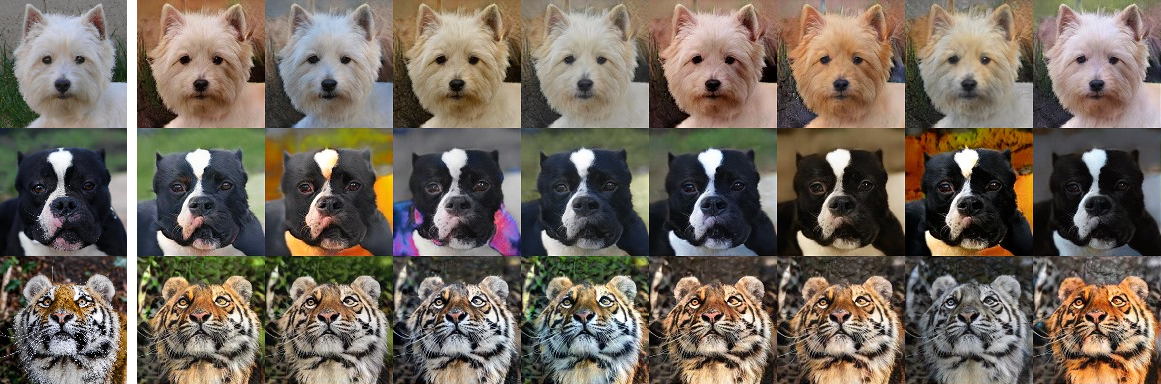}
	\caption{Animal-to-animal translation results with $16\times 16 \times 128$ intermediate features.}
	\label{fig:a2a16x16x128}
\end{figure*}
% We use FFHQ and AFHQ datasets for this experiment. 
The results of human face transfer are shown in Fig.~\ref{fig:f2f16x16x128}. The first column is the input image and the following columns are the corresponding output images. The output images look like a series of selfies of similar people with different detail texture. The global topology information, such as the positions of the eyes, nose, and mouth, is maintained in the same positions as the input. The detail features, such as skin folds and hair color, are randomly set.

The same model was been applied to the AFHQ dataset. However, the number of images is only $3000$, while the human face dataset has $70000$ images. The result is shown in Fig.~\ref{fig:a2a16x16x128}. Compared to Fig.~\ref{fig:f2f16x16x128}, the only difference is the colors of the output images in the same row. All other features remain the same.

The varying outcomes of the two experiments are due to differences in the sizes of the datasets. When the size of the dataset is relatively low, the discriminator possesses sufficient capability to distinguish differences, causing the output the GAN converge that of autoencoder. This demonstrates the validity of the analysis in the method section.

\subsubsection{Image Translation with Low Dimension Features Transfer}
\begin{figure*}
	\centering
	\includegraphics[width=16cm]{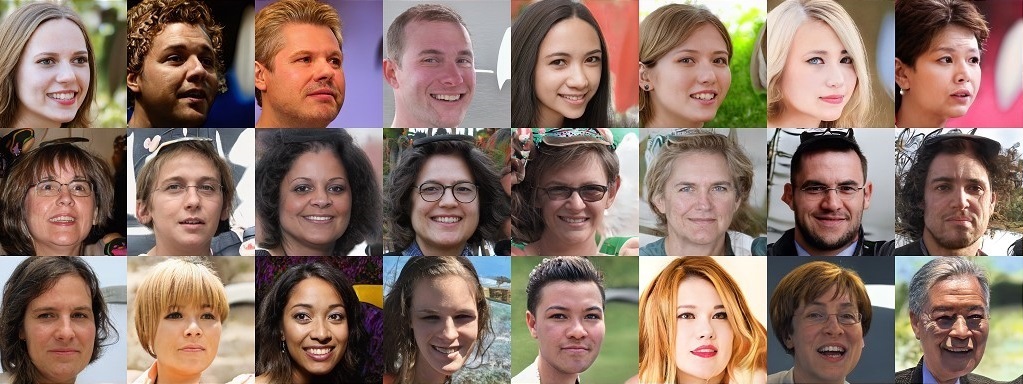}
	\caption{Face-to-face translation result with $8\times 8 \times 128$ intermediate feature. }
	\label{fig:f2f8x8x128}
\end{figure*}

\begin{figure*}
	\centering
	\includegraphics[width=16cm]{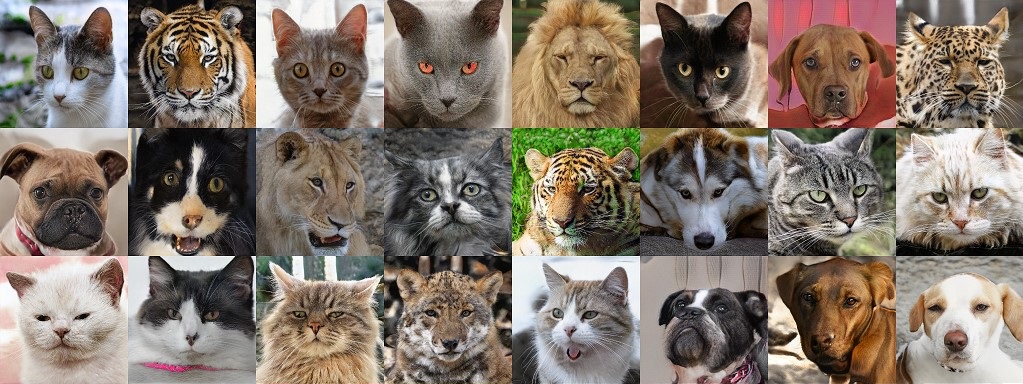}
	\caption{Animal-to-animal translation results with $8\times 8 \times 128$ intermediate features.}
	\label{fig:a2a8x8x128}
\end{figure*}
% The same self to self-transfer has been applied with lower-dimension features.
The experiment in this subsection is similar to previous subsection. The only difference is the intermediate feature decrease from $16\times 16 \times128$ to $8\times 8 \times 128$, which makes the encoder not able to reserve all features from inputs. The result is shown in Fig.~\ref{fig:f2f8x8x128} and~\ref{fig:a2a8x8x128}.

In face-to-face translation, the input image is in the first column, followed by the corresponding output images in the subsequent columns. Unlike in Fig.~\ref{fig:f2f16x16x128}, the difference between each image in the same row is more pronounced. The image does not depict people with slightly different. Instead, Fig.~\ref{fig:f2f8x8x128} shows people of different sex, gender, and other details. The common feature is that they take selfies from the same angle and maintain the same pose.

In Fig.~\ref{fig:a2a8x8x128}, discerning the similarity becomes even more challenging. %The input is shown in the first column, while the rest display the output.
The first column shows the input, while the rest display the output.
It shows that within the same row, the animal species and angles of the photos differ. However, we observed that, at the beginning of the training process, the network retains the pose and angle of the input image for animal data. However, as training progresses, these features are discarded to enhance the realism of the output image if the capacity of the network is not insufficient. This is because the network is confused on which part of the feature should be preserved. This also shows that our analyses are correct.

% \textcolor{red}{The network not sure which part of common feature should be preserved}

% This result reveals varying outcomes across two distinct datasets.

% This suggests that insufficient dataset size to encompass the entire space causes the network to lose global topology features.

% This experiment shows that the network will not able to preserver most
% information if we compress the feature to event lower dimension. The network can
% only keep the global topology information.

\label{exp1}
\section{Conclusion}
Our study provides new insights into the effectiveness of GANs in tasks involving image-to-image translation. We have shown that adversarial training, when applied to autoencoder models, can achieve results comparable to traditional methods without the necessity for additional complex loss penalties. Furthermore, we explained the differences and similarities between GANs and autoencoders. We also incorporated experimental results to demonstrate the validity of our findings.

\ifCLASSOPTIONcaptionsoff
  \newpage
\fi




\bibliographystyle{IEEEtran}
% % argument is your BibTeX string definitions and bibliography database(s)
\bibliography{IEEEexample}
\end{document}